\documentclass[11pt,a4paper]{article}

\usepackage[]{naacl2021}
\usepackage{times}
\usepackage{latexsym}
\usepackage{amsmath}
\usepackage{multirow}
\usepackage{graphicx}
\usepackage{enumitem}
\usepackage{subcaption}
\usepackage{caption}
\usepackage{booktabs}

\usepackage{xcolor}

\newcommand{\prism}{\newcite{thompson20}}

\usepackage{microtype}

\title{Assessing Reference-Free Peer Evaluation for Machine Translation}

\author{Sweta Agrawal$^1$, George Foster$^2$, Markus Freitag$^2$, Colin Cherry$^2$ \\
  $^1$Department of Computer Science, University of Maryland \\
  $^2$Google Research \\
  \texttt{sweagraw@umd.edu}, \texttt{\{fosterg, freitag, colincherry\}@google.com} \\}

\date{}

\begin{document}
\maketitle
\begin{abstract}
%
Reference-free evaluation has the potential to make machine translation evaluation substantially more scalable, allowing us to pivot easily to new languages or domains. It has been recently shown that the probabilities given by a large, multilingual model can achieve state of the art results when used as a reference-free metric.
We experiment with various modifications to this model, and demonstrate that by scaling it up we can match the performance of BLEU. We analyze various potential weaknesses of the approach, and find that it is surprisingly robust and likely to offer reasonable performance across a broad spectrum of domains and different system qualities.

\end{abstract}

\section{Introduction}

Traditional automatic metrics for machine translation (MT), such as BLEU~\cite{papineni-etal-2002-bleu}, score MT output by comparing it to one or more reference translations. This has several disadvantages. First, high-quality reference translations are expensive to create. This means that in practice, evaluation is usually carried out with relatively small, carefully curated test corpora. The need for careful preparation limits the number of domains for which an MT system can be conveniently assessed, and small test-set sizes can make it difficult to draw robust conclusions \cite{card2020little}. Second, enshrining ground truth in a small number of references (usually just one) is inherently problematic, since valid translations can vary along many dimensions; \newcite{freitag-etal-2020-bleu} demonstrate that different (correct) references for the same test set can result in different system rankings according to the same reference-based metric.
Finally, scoring the similarity between an MT hypothesis and a reference translation involves recognizing the extent to which they are mutual paraphrases. When gross discrepancies exist, this is a relatively easy problem for which surface-level metrics can provide a reliable signal, but capturing the subtle errors typical of high-quality MT is more difficult, and it is not clear whether it is substantially easier than scoring the similarity between texts in different languages.

These problems can be avoided by looking only at the source text when assessing MT output. There is evidence that this is the best practice for human evaluation
\cite{toral-2020-reassessing}. Moreover, it has recently been investigated for automatic metrics as well \cite{yankovskaya-etal-2019-quality,lo-2019-yisi,zhao2020limitations,ma-etal-2019}.
Such {\em reference-free} metrics are flexible and scalable, but since they are essentially performing the same task as an MT model,
they raise a circularity concern: if we can reliably score MT output, why wouldn't we use the scoring model to produce better output? One answer to this is practical: the scoring model might be too large to deploy, or it might not easily support efficient inference \cite{yu2016neural}. 
A more interesting answer is that a scoring model could be set up to provide a signal that is complementary to the systems under evaluation. That is, it might be capable of correctly ranking competing MT hypotheses even when its own preferred hypothesis is worse on average than those of the systems it is evaluating. In our experiments we find that this can indeed be the case.

In recent work, \newcite{thompson20} showed that a single multilingual MT model trained on 39 languages can achieve excellent paraphrase recognition when used in zero-shot mode to compare MT output with reference sentences in the same language. On the WMT 2019 metrics task,
their method (Prism) beat or tied all previous reference-based metrics on all languages.\footnote{Except Gujarati, which was absent from their training corpus.} Although it was not the main focus of their work, Prism  achieved a new state-of-the-art as a reference-free metric, simply scoring target given source text using an MT model, in a post-competition comparison to  the 2019 ``Quality Estimation as a metric'' shared task \cite{ma-etal-2019}.

Our aim in this paper is to characterize the conditions under which the Prism approach---using one MT system to perform {\em peer evaluation} on other systems---can be successful: what properties does the evaluating system need to have, how powerful should it be, and how close can it be to the systems under evaluation?
We focus on system-level evaluation, which we believe is the most compelling use case for reference-free methods, targeting a broad characterization that complements the potentially more precise picture furnished by reference-based metrics for a specific test corpus.
We first replicate the correlation with human judgment results from \newcite{thompson20} on WMT 2019, using the same corpora and architecture. Next, we examine several alternative design decisions in an attempt to improve Prism and further our understanding. These include the effects of varying training corpora (domain, number of languages, use of monolingual data); model capacity (scaling up and down from the original architecture); and different methods for regularizing token-level probabilities (Monte-Carlo dropout, subword sampling) and for combining them into system-level scores (summary statistics over tokens, confidence thresholds over sentences). Finally, we analyze the results of our best model, measuring how its performance depends on various factors: language pair and human-judgment methodology, output quality, proximity to the systems under evaluation, and size of the test set. 

We demonstrate improvements over the original Prism metric due to model capacity and different methods for combining probabilities; surprisingly, we find little gain from adjusting the domain or languages in the original multilingual corpus (although we show that a competition-grade English-German system outperforms the generic multilingual system). We find that the evaluating MT system's output quality is generally correlated with its performance as a metric, although we corroborate the surprising finding from \newcite{thompson20} that it is not necessary to be the best---our system is middle-of-the-road or worse according to BLEU across most WMT 2019 languages. We measure the proximity between our system and the systems under evaluation and find no evidence that this is a source of bias. 
Despite using no references, our model achieves approximate parity with BLEU both in system-level correlation with human judgment, and when used for pairwise comparisons.

\section{Related Work}

Reference-free evaluation is widely used for many NLP tasks such as grammatical error correction \cite{napoles2016there}, dialog \cite{sinha2020learning,mehri2020usr} and text generation \cite{ethayarajh2020bleu}. There has been recent interest in reference-free evaluation for MT, which was a joint track between the WMT 2019 metrics task \cite{ma-etal-2019} and quality estimation task \cite{fonseca-etal-2019-findings}. Reference-free metrics competed head-to-head with standard metrics, and generally did worse. However, the results from the best reference-free systems, UNI+ \cite{yankovskaya-etal-2019-quality} and YiSi-2 \cite{lo-2019-yisi} were surprisingly close to the standard metric scores on the language pairs for which they were evaluated.



UNI+ computes word-level embeddings for source and MT output sentences using pre-trained multilingual BERT and LASER \cite{artetxe-schwenk-2019-margin} models, then feeds averaged vectors to a neural classifier trained to predict human scores from previous MT metrics tasks. YiSi-2 is similar, except that it works in an unsupervised fashion, computing similarities between mBERT embeddings for aligned source and target words, and returning an F-measure statistic. 
In more recent work, \newcite{zhao2020limitations} adopt a similar approach based on mBERT, aligning representations from multilingual embedding spaces before computing distances with MoverScore \cite{Zhao2019MoverScoreTG}, and adding a GPT-based target-side language model.
The current state-of-the-art in reference-free evaluation for MT is represented by the Prism approach \cite{thompson20} which we extend here.

It is worth distinguishing reference-free evaluation from two related tasks that share formal similarities. The first is quality or confidence estimation \cite{blatz-etal-2004-confidence,Specia2018,chelba2020data}, which aims to score the fitness of MT output for a downstream application. This is typically supervised, although a recent approach \cite{Fomicheva2020UnsupervisedQE} dispenses with the need to learn from human annotations, as do most of the approaches we study in this paper. Quality estimation is most usefully applied at the sentence level, and it can make use of powerful ``glass-box'' features which capture the internals of an MT system. In contrast, reference-free evaluation is most naturally applied at the system (test-set) level, and ideally should make no assumptions about the systems under evaluation. The second task is parallel-corpus mining \cite{Zhang2020ParallelCF,Yang2019ImprovingMS}, which aims to identify valid translations at various levels of granularity. Its scoring aspect is similar to reference-free evaluation, but it is applied to a different input distribution, attempting to identify human-generated translation pairs rather than scoring MT outputs for a given human-generated source text.

\section{Methods}

We aim to generate a quality score $s(X,Y) = \sum_{x,y} s(x,y)$ for source and target texts $X,Y$ which consist of segment (nominally, sentence) pairs $x,y$. We assume no document or ordering information among segments, and do not directly evaluate scores for individual segment pairs.
All methods we consider make use of token-level log-probabilities from a standard autoregressive neural MT system: $\log p(y_t|y_{<t}, x)$, where $y = y_1\ldots y_{T}$. We experimented with reverse probabilities $p(x|y)$, but like \newcite{thompson20} found these gave no advantage, and do not include them in our reported results. The following sections describe our model architecture, scoring techniques, and evaluation methodology.

\subsection{Model}

Our baseline NMT model uses a standard Transformer architecture identical to that of \newcite{thompson20} (up to toolkit differences), trained on the same multilingual corpus.
To encourage language-agnostic encoder representations for zero-shot scoring, the baseline uses target-language tags at the beginning of each target sentence \cite{johnson2017google}. Since we do not require such representations for reference-free evaluation, we also tried introducing the tags earlier, at the beginning of each source sentence. We vary training corpora and model capacity as described in section~\ref{sec:data}, but otherwise make no changes to the model.

\subsection{Scoring} \label{sec:scoring}

We investigated various techniques for deriving segment-level scores $s(x, y)$: regularization, different methods for aggregating token-level probabilities, and segment-level confidence thresholds.

\subsubsection*{Regularization}

To obtain smoother scores, we used Monte-Carlo dropout \cite{gal2016dropout} and subword regularization \cite{kudo2018subword}. These involve estimates of the form:
\[
\log p(y|x) = \sum_{k=1}^K \log p_k(y|x) / K,
\]
where $p_k(y|x)$ is a probability estimate that depends on the smoothing method.
For MC-dropout, it is obtained by dropping neural connections with probability $\alpha$.
For subword regularization, $p_k(y|x) = p(\tilde{y}_k|\tilde{x}_k)$, where $\tilde{x}_k$ and $\tilde{y}_k$ are randomly-sampled alternative subword segmentations of $x$ and $y$.\footnote{We perform an approximate search for the 10-best subword segmentations, then sample from this list with probability proportional to a unigram estimate  $q^\alpha(\tilde{x}|x)$.} Note that MC-dropout decomposes over tokens, yielding smoother per-token probabilities; subword regularization does not, since it does not preserve tokenization.

\subsubsection*{Aggregating token-level log-probabilities}

Given a sequence of token probabilities $\log p(y_t|y_{<t}, x)$, $t=1\ldots T$, we derive segment-level scores $s(x, y)$ using various statistics. Following \prism, we sum to obtain segment log-probabilities or average to obtain mean token-wise log-probabilities. To eliminate the effect of outliers, we tried the median instead of the mean. To test the opposite intuition, we also tried the minimum.
Finally, to reflect overall consistency, we compute standard deviation.

\subsubsection*{Confidence Thresholds}

Quality scores implicitly reflect the presence or absence of errors in MT output. In some cases, model probabilities provide strong evidence for or against the existence of errors, but in other cases the model may be agnostic. To capture this intuition, we used the following mapping to obtain segment scores:
\[
s(x, y) = \left\{
    \begin{array}{ll}
    -1, & \log p(y|x)/T < l \\
    +1,  & \log p(y|x)/T > h \\
    0,  & \mbox{else}
    \end{array}
\right.         
\]
To set the thresholds $(l, h)$ we used a coarse grid search on development data.


\subsection{Evaluation} \label{sec:eval}

We evaluate reference-free metric scores on data from the WMT19 metrics task \cite{ma-etal-2019}, consisting of outputs from different MT systems for 18 language pairs. For each language pair, we compute a metric score for each system, then use correlation with the provided human scores to assess the quality of our metric.\footnote{Human annotators assign segment-level scores on a $0-100$ scale which are averaged across segments, then normalized to correct for annotator differences, then averaged across annotators to produce system-level scores. For out-of-English language pairs, annotations are made by comparison to the source text, which directly corresponds to our setting; for other pairs, they are made by comparing to reference translations.}
Following \newcite{ma-etal-2019} we measure correlation using Pearson’s coefficient, and use Williams' test \cite{williams59} to compute the significance of correlation differences, with a p-value $<$ 0.05.

\newcite{ma-etal-2019} note that correlation scores are unrealistically high for many language pairs, and suggest using only the best $k$ systems for small values of $k$. However, \newcite{mathur20} show that this results in noisy and unreliable estimates. We adopt their suggestion to instead remove outlier systems whose scores have large deviations from the median according to the formula:
\[
\frac{|h - \tilde{h}|}{1.483 \times  \mbox{median}_h(|h - \tilde{h}|)} > 2.5,
\]
where $h$ is a system-level human score, and $\tilde{h}$ is the median score across all systems for a given language pair.

To summarize a metric's performance across a set of language pairs, we report the weighted average of its Pearson correlations across languages. We first apply the Fisher Z-transformation to normalize raw language-specific correlations, then weight by the number of MT systems per language (post outlier filtering), then invert the Fisher Z-transformation and take the mean \cite{hedges2014statistical}.

\section{Experimental Settings}

\subsection{Data} \label{sec:data}

We used four training corpora. {\bf Prism-39} consists of noise-filtered multi-way parallel data curated by \prism, extracted primarily from Wikimatrix, Global Voices, EuroParl, SETimes, and United Nations, consisting of 99.8M sentence pairs in 39 languages, including direct parallel data for 706 language pairs. {\bf Wiki-39-Mono} consists of monolingual data extracted from the multilingual Wikipedia corpus for the languages available in Prism-39. {\bf WMT-15} is the parallel training data provided for the WMT 2019 News Translation Task \cite{barrault2019findings}, augmented with 5 languages from previous WMT years---Estonian (et), Spanish (es), Latvian (lt), Hindi (hi) and Turkish (tr). All language pairs are to/from English except French-German. Sizes range from 60 million sentence pairs for English-Czech to 10k pairs for English-Gujarati (Table~\ref{tab:data1}). Finally, {\bf WMT-15-Mono} is the monolingual data provided alongside WMT-15.

Test data is from the WMT 2019 Metrics Task \cite{ma-etal-2019}, consisting of system outputs on news-domain text for all 18 language pairs included in the task: English (en) to/from  Czech (cs), German (de), Finnish (fi), Gujarati (gu), Kazakh (kk), Lithuanian (lt), Russian (ru), and Chinese (zh), excluding cs-en. There are three other language pairs not including English: de-cs, de-fr and fr-de. The average number of systems per language is 12, and the average test-set size is 1,633.


\subsection{MT Systems}

\begin{table}[!htp]\centering
\scalebox{0.8}{
\begin{tabular}{lrrrrrr}\toprule
 \textbf{Scale}  & \textbf{Params} & \textbf{Layers} & \textbf{Hidden} & \textbf{Heads}  & \textbf{Model}\\\midrule 
 Big &  473M &6 & 8192 & 16 & 1024 \\ 
 Prism & 900M & 8 & 12288 & 20 & 1280 \\ 
 Massive & 1.8B & 8 & 16384 & 32 & 2048 \\ 
\bottomrule
\end{tabular}}
\caption{Model configurations used in our experiments.} \label{tab:modelconfig}
\end{table}

We used the Lingvo toolkit \cite{shen2019lingvo}, to train Transformer sequence-to-sequence models of various sizes as shown in Table~\ref{tab:modelconfig}, where the baseline {\em Prism} configuration matches that of \prism. We use AdaFactor optimization with a learning rate of 1.0 and batch size of $\sim$8000 samples. Our shared vocabulary comprises 64k subwords.

\section{Results}

This section presents our main results. All correlations in the tables below are for system-level scores, after outlier systems have been discarded for each language pair.
For brevity, we report average correlations, normalized and weighted as described in section~\ref{sec:eval}; full results are provided in Appendix~\ref{sec:results-full}. Unless otherwise stated, all methods score system outputs using average log probabilities normalized by segment length.

\subsection{Baselines}

\begin{table}[!htp] \centering
\scalebox{0.85}{
\begin{tabular}{lrrrr}\toprule
\textbf{Metric} & All & en-xx & xx-en & xx-yy \\\midrule
BLEU & 0.911 & 0.917 & 0.921 & 0.838	 \\
CHRF & 0.933  & 0.937 & 0.919 & 0.954 \\
\midrule
UNI+\text{*} & 0.808 & 0.746 & 0.822 & - \\
Yisi-2\text{*} & 0.487 & 0.272 & 0.646 & 0.489 \\
Prism & {\bf 0.861} & 0.814 & {\bf 0.887} & 0.911\\
\midrule
Prism-trg2xx & 0.853 & 0.812 & 0.872 & 0.907\\
Prism-src2xx & 0.858 & {\bf  0.822} & 0.871 & {\bf 0.914}\\
\bottomrule
\end{tabular}}
\caption{Baseline results. All numbers are average system-level correlations. \text{*}Average is computed over language pairs for which the corresponding metric had a submission in the WMT19 Metrics task. } \label{tab:baselines-small}
\end{table}

Table~\ref{tab:baselines-small} shows key WMT19 baseline results for reference-based metrics (top two lines), reference-free metrics (next three lines), and our reimplementation of the Prism model (bottom lines). We achieve slightly better results for source-side tagging (Prism-src2xx), and on average match the original Prism results that use target-side tagging with this configuration, which we adopt for further experiments.
The {\em en-xx} results are affected negatively by the inclusion of en-gu, which is absent from the Prism-39 corpus and has low correlation ($0.400$); however, interestingly, results for gu-en are on par with other language pairs, presumably due to the prevalence of English in the corpus.

\subsection{Training data}

\begin{table}[!htp] \centering
\scalebox{0.75}{
\begin{tabular}{lrrrr}\toprule
\textbf{Data} & All & en-xx & xx-en & xx-yy \\\midrule
Prism-39 & 0.858 &  0.822 & 0.871 & 0.914 \\
WMT-15 & 0.840 & 0.776 & \underline{0.890} & 0.854 \\
Prism-13 & 0.863 & 0.828 & \underline{0.888} & 0.888\\
Prism-39 + WMT-15 & \underline{0.867} & 0.811 & \underline{0.896} & \underline{{\bf 0.923}} \\
\midrule
\textit{Adding monolingual data} \\
Prism-39 + Wiki-39  & 0.832 & 0.792 & 0.859 & 0.869 \\
WMT-15 + WMT-15-Mono &  {\bf 0.870} & {\bf 0.839} & \underline{{\bf 0.910}} & 0.818 \\
Prism-39 + WMT-15-Mono & 0.851 & 0.831 & 0.863 & 0.874 \\
\bottomrule
\end{tabular}}
\caption{Effect of training data. Significant improvement over baseline ``Prism-39'' systems are underlined. 
} \label{tab:corpora-small}
\end{table}

Table~\ref{tab:corpora-small} gives results for training on different corpora described in section~\ref{sec:data}.
The first four lines correspond to different multilingual training corpora, beginning with the Prism-39 model from the previous section. We see no gain on average from using the provided WMT-15 training corpora, despite possibly better domain fit and generally larger sizes for the language pairs in the test set (Table~\ref{tab:data1}). We speculate that this is due to preprocessing as we made no effort to clean or filter the WMT-15 corpus. This hypothesis is supported by the  Prism-13 results, where we trained on the language pairs in Prism-39 that overlapped with the WMT-15 corpus, achieving slightly better average performance. Combining Prism-39 and WMT-15 improves further, yielding a relatively small but statistically significant average gain over pure Prism-39, at the cost of lower performance for the en-xx language pairs.

\begin{table}[!htp]\centering
\scalebox{0.95}{
\begin{tabular}{lrr}\toprule
\textbf{LP} & \textbf{Prism-39} & \textbf{WMT-15} \\
\midrule
en-zh &	1.49 & 64.33\\
en-fr &	3.52 & 40.44\\
en-ru &	2.25 & 38.49\\
en-cs &	0.65 & 25.98\\
en-es &	4.40 & 15.18\\
de-fr &	0.65 & 9.82\\
en-fi &	0.28 & 6.58\\
en-de & 1.36 & 4.50\\
en-et &	0.22 & 2.17\\
en-lv &	0.09 & 0.63\\
en-lt & 0.16 & 0.63\\
en-kk & 0.20 & 0.22\\
en-gu & 0.00 & 0.01\\
de-cs &	0.37 & 0.00\\
\bottomrule
\end{tabular}}
\caption{Corpus size for overlapping language pairs from Prism-39 and WMT-15 (in millions of segments): WMT-15 has more parallel data available for all languages except de-cs, where no parallel corpora is available in the WMT-15 dataset.} \label{tab:data1}
\end{table}

\begin{table*}[!htp]\centering
\scalebox{0.80}{
\begin{tabular}{lrrrr|rrrr}\toprule
\textbf{Data} &en-de & en-lt & en-ru & en-zh  & de-en& lt-en & ru-en & zh-en \\\midrule
BLEU & 0.806 & 0.986 & 0.946 & 0.802 & 0.794 & 0.985 & 0.812 & 0.808 \\
\midrule
Prism-39 &  0.730 & \underline{\bf 0.939} & {\bf 0.901} & {\bf 0.789} &  0.796 &	\underline{\bf 0.978} &	{\bf 0.739} &	{\bf 0.828}\\
Bilingual Models  &0.726 & 0.695 & 0.867 & 0.769& {\bf 0.801} & 0.862 & 0.650 & 0.826 \\
\midrule
Competition-grade &  {\bf 0.913} & - & -& -& -& -& -& -\\
\bottomrule
\end{tabular}
}
\caption{Bilingual vs multilingual models for scoring.} \label{tab:bivsmulti}
\end{table*}

\begin{table}[!htp]\centering
\scalebox{0.95}{
\begin{tabular}{lrr}\toprule
\textbf{LP} & \textbf{WMT-15-Mono} & \textbf{Wiki-39} \\
\midrule
de & 275.69 & 59.93\\
en & 199.90 & 130.79\\
fr & 160.93 & 48.52\\
lt & 106.19 & 4.85\\
ru & 80.14 & 46.27\\
cs & 72.15 & 9.44\\
fi & 18.84 & 7.88\\
kk & 13.82 & 3.34\\
gu & 4.64 & 0.00\\
zh & 2.15 & 21.79\\
\hline
et & 51.68 & 3.03\\
es & 43.81 & 36.17\\
hi & 23.61 & 2.37\\
lv & 10.20 & 1.36\\
tr & 9.65 & 5.42\\
\bottomrule
\end{tabular}}
\caption{Corpus size for overlapping languages from WMT-15-Mono and Wiki-39 dataset (in millions). The last five languages are not a part of WMT'19 Metrics evaluation task but were included when training the multilingual MT system. } \label{tab:data2}
\end{table}

Inspired by improvements for low-resource languages from monolingual data \cite{siddhant-etal-2020-leveraging}, we used the MASS denoising objective
to add general-domain monolingual data (Wiki-39) to Prism-39 and in-domain data (WMT-15-Mono) to both Prism-39 and WMT-15 (Table~\ref{tab:data2} for a comparison on the relative sizes of the monolingual corpora). Overall, the general-domain data hurts correlation significantly, while in-domain helps significantly, but only for WMT-15. As expected, monolingual data tends to help lower-resource languages (gu, kk, lt) most, with a particularly large gain for {\em xx-en } with WMT-15 + WMT-15-Mono. However, the correlation for {\em xx-yy} language-pairs degrades significantly, which we attribute to the en-centric nature of the WMT-15 dataset.


\subsection{Bilingual Systems}

Can we use bilingual MT systems for peer evaluation? We chose four representative language pairs from Prism-39 and trained ``Big'' models (see Table~\ref{tab:modelconfig}) in eight directions, with dedicated 64k subword vocabularies. Table~\ref{tab:bivsmulti} shows that for medium and high resource languages (de, ru, and zh), the bilingual model performs comparably to the multilingual model. However, for the low resource language ``lt'', the multilingual model is significantly better.
As with the results elsewhere in this section, this suggests that correlation tends to follow the pattern one would expect if we were mainly interested in model quality. This is corroborated by the results in the last line of the table, where we compare a competition-grade model for en-de \cite{Freitag20a}, similar to the winning submission from WMT19, to our models. The competition-grade model achieves a much better correlation and also improves on BLEU by a wide margin.



\subsection{Model Capacity}

\begin{table}[!htp] \centering
\scalebox{0.90}{
\begin{tabular}{lrrrr}\toprule
\textbf{Metric} & All & en-xx & xx-en & xx-yy \\\midrule
BLEU &  0.911  &  0.917 &  0.921 & 0.838 \\
\midrule
Big & 0.808 & 0.745 & 0.838 & 0.885\\
Prism & 0.858  &  0.822 & 0.871 & 0.914 \\
Massive & {\bf 0.883} & {\bf 0.858} & {\bf 0.890} & {\bf 0.927} \\
\bottomrule
\end{tabular}}
\caption{Effect of Model capacity.
} \label{tab:capacity-small}
\end{table}

Motivated by the link between correlation and model quality, we varied model capacity according to the settings in Table~\ref{tab:modelconfig}, using the Prism-39 training corpus. The results in Table~\ref{tab:capacity-small} show a clear pattern of gains with increasing capacity. The {\em Massive} configuration does best overall,
achieving  statistical parity with BLEU on average.

\subsection{Scoring Methods}

\begin{table}[!htp]\centering
\scalebox{0.80}{
\begin{tabular}{lrrrr}\toprule
\textbf{Aggregation Method} & All & en-xx & xx-en & xx-yy\\\midrule
Mean & 0.883 & 0.858 & 0.890 & {\bf 0.927}  \\
Std-dev &  0.882 & 0.847 & {\bf 0.903} & 0.919 \\
Median & 0.870 & 0.859 & 0.876 & 0.887\\
Min  & 0.872 & 0.840 & 0.895 & 0.896\\
\midrule
MC-dropout (Mean) & 0.877 & 0.847 & 0.888 & 0.926	\\
SP-Norm (Mean) &  0.884 & 0.861 & 0.892 & 0.924\\
\midrule
Confidence threshold & {\bf 0.886} & \underline{{\bf 0.898}} & 0.858 & 0.910 \\
\bottomrule
\end{tabular}}
\caption{Scoring methods. Significant improvements over baseline {\em Mean} systems are underlined.} \label{tab:alternate-small}
\end{table}

Table~\ref{tab:alternate-small} shows results for the scoring methods described in section~\ref{sec:scoring} applied to the Massive configuration. Aggregating token probabilities using statistics other than mean gives small gains on some languages, but hurts on average. Regularizing with MC-dropout or subwords (SP-norm) leads to significant gains in some cases, with a slight overall increase over mean for SP-norm. We tuned confidence thresholds on WMT18 Metrics task data using a grid of 16 log-probability points in $[-3, 0]$, which yielded optimal thresholds $(-1, -0.6)$. This produced our best overall result, with systematic gains on {\em en-xx} pairs.  


\section{Analysis}

In this section we analyze various aspects of metric performance, confining our attention to the Massive model with mean scoring for consistency.

\subsection{Performance across conditions}

\begin{table}[!htp]\centering
\scalebox{0.9}{
\begin{tabular}{lr}\toprule
 \textbf{Subset} &  \textbf{Avg} \\\midrule
 All & 0.883 \\
 All - gu & 0.893 \\
 \midrule
 Source-based evaluation &  0.858 \\
 Source-based - gu & 0.883 \\
 Reference-based evaluation & 0.901 \\
 Reference-based - gu & 0.901\\
 \midrule
 Corpus $\ge$1M & 0.839\\
 Corpus $<$1M & 0.924 \\
 No data & 0.741 \\
\bottomrule
\end{tabular}}
\caption{Average Correlation for different subsets of languages.} \label{tab:subset}
\end{table}

Different languages have different relations to our model, to the systems participating in the WMT task, and to the human scoring procedure used in the WMT19  data. Table~\ref{tab:subset} shows results for various conditions. Removing the language (gu) for which we have no training data improves average correlation substantially. 
The human evaluations for out-of-English language pairs involve comparing MT output to the source text; the evaluations for remaining pairs involve comparing it to reference translations.  
We see no boost from the language pairs for which source-based human evaluation was used (matching our setting), and in fact do somewhat worse on these pairs than the others, on average. Finally, we achieve better performance for lower-resource ($<$ 1M parallel segments) language pairs than higher-resource pairs 
(with respect to the Prism-39 corpora), but poor average performance on the pairs (en-gu/gu-en) for which we had no training data.


\subsection{Pairwise comparisons}

Correlation statistics give an overall picture of metric performance, but do not directly reflect the frequent use case of deciding which of two systems is better. To measure this, we examined whether our metric agrees with human pairwise ranking decisions over all pairs of systems. Following \cite{mathur20}, we apply the Wilcoxon ranksum test and paired t-test to detect when such decisions are significant according to human and metric scores respectively.

\begin{table}[!htp]\centering
\scalebox{0.80}{
\begin{tabular}{lcccccc}\toprule
\textbf{Metric} & \multicolumn{3}{c}{\textbf{Human-S}}& \multicolumn{3}{c}{\textbf{Human-NS}} \\ 
& C ($\uparrow$) & IC ($\downarrow$)  & NS & C ($\uparrow$) & IC ($\downarrow$) & NS \\
\midrule
\textit{All Systems} \\
BLEU & 768  & { \bf 37} & 80 & 126 & { \bf 71} & 70\\
Prism & { \bf 778} & 61 & 46 & { \bf 136}  & 93 & 38\\
\textit{en-xx Systems} \\
BLEU & 411 & { \bf 25} & 53 & 38 & { \bf 25} & 26\\
Prism & { \bf 421} & 36 & 32 &  { \bf 39} & 29 & 21\\
\textit{xx-en Systems} \\
    BLEU  & { \bf 285} & { \bf 8} & 20 & 67 & { \bf 40} & 33\\
Prism & 277 & 23 & 13 & { \bf 74} & 53 & 13 \\
\textit{xx-yy Systems} \\
BLEU & 72 & { \bf 4} & 7 & 21 & { \bf 6} & 11\\
Prism & { \bf 80} & 2 & 1 & { \bf 23} & 11 & 4\\
\bottomrule 
\end{tabular}
}
\caption{WMT19 pairwise system level comparisons using the Massive configuration: Human-NS and Human-S means insignificant and significant differences according to human scores; C and IC stands for Correct and Incorrect ranking according to metric and human scores; NS represents insignificant differences according to the metric scores.} \label{tab:pairwise}
\end{table}

Table~\ref{tab:pairwise} shows ranking performance for Prism compared to BLEU, categorized according to language pair grouping. The general pattern across all groupings is that Prism is more decisive: it makes more significant decisions than BLEU, leading to higher rates of both correct and incorrect rankings.
Among the 885 system pairs (across all languages) that are considered significantly different according to human judgment, Prism correctly ranks 88\% with significantly different scores, compared to 87\% for BLEU.



\subsection{Quality of the evaluating model}

\begin{figure}[htp]
    \centering
    \includegraphics[width=0.45\textwidth]{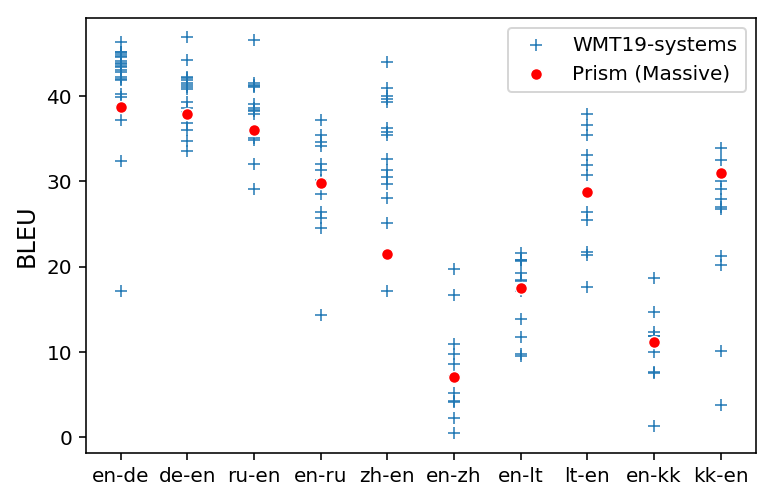}
    \caption{Quality across language pairs. }
    \label{fig:quality}
\end{figure}

How good is our multilingual MT system compared to the systems under evaluation?
We generated translations of the test text for a subset of languages and compared the quality of the generated system outputs using BLEU.
Figure~\ref{fig:quality} shows that our evaluating model achieves worse BLEU scores than many of the systems under evaluation, ranking around the median for most language pairs. Although Table~\ref{tab:bivsmulti} provides evidence that stronger systems produce better metrics, clearly it is not necessary to be among the top-ranked systems in order to generate a signal that is approximately as reliable as BLEU.\footnote{It would be interesting to try to characterize the relation between system quality and metric strength more precisely, but in the absence of human judgments of our output quality, any such picture we could currently draw would be clouded by metric noise.}

\subsection{Proximity Bias}

A potential pitfall in peer evaluation is bias toward one or more of the systems under evaluation. Clearly, the evaluating system will prefer its own output---how far from an evaluated system does it have to be in order to judge it fairly? Lacking access to the systems in the WMT19 task, we measure proximity using cross-BLEU score (using one output as hypothesis and the other one as reference translation) between the system output and the output generated by our Prism model. In the presence of bias, we would expect the metric to result in higher ranking for closer systems and lower ranking for farther systems (relative to human scores).

\begin{figure}[htp]
    \centering
    \includegraphics[width=0.45\textwidth]{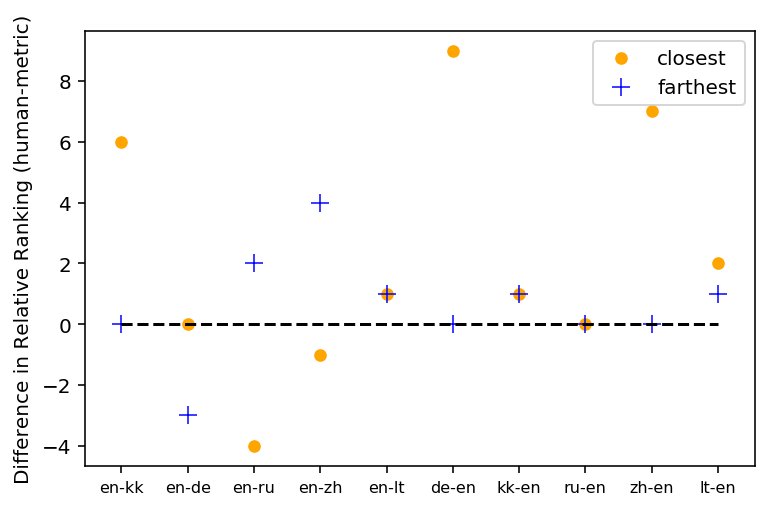}
    \caption{Relative Ranking of the closest and farthest systems under evaluation to the Prism system as measured by cross-BLEU.}
    \label{fig:cross-bleu}
\end{figure}

\begin{figure}[htp]
    \centering
    \includegraphics[width=0.45\textwidth]{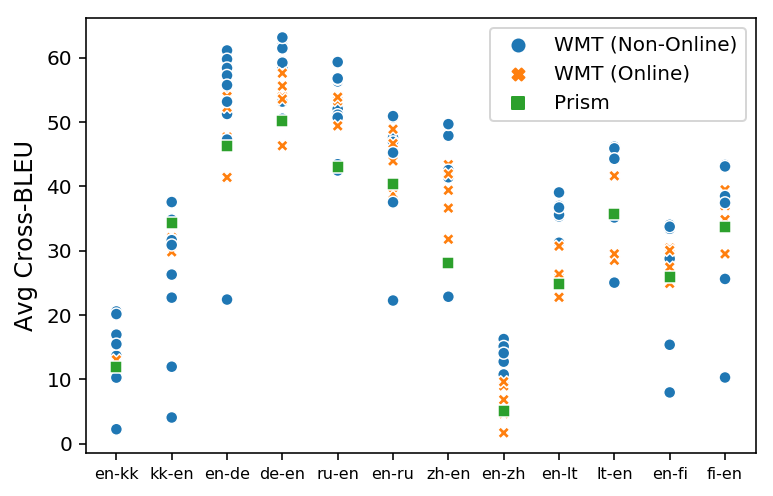}
    \caption{Average Cross-BLEU for all evaluated systems and Prism.}
    \label{fig:avg-cross-bleu-all}
\end{figure}

Figure~\ref{fig:cross-bleu} shows the relative ranking of the closest and the farthest system to Prism (relative to human).
Since the model makes mistakes in both directions---ranks closest and farthest system both higher and lower than human---there is no evidence from this analysis that it exhibits a strong bias in favour of systems whose outputs are closer to its own. A potential explanation is that it is sufficiently far from most of the evaluated systems due to its multilingual training corpus. To verify this, we computed the average cross-BLEU for each evaluated system (relative to all others), and compared it to the same quantity for our system. Figure~\ref{fig:avg-cross-bleu-all} shows that we are indeed an outlier system for most language pairs. The systems with lower cross-BLEU than Prism are mostly online or rule-based systems.\footnote{For Kazakh (kk), Prism-39 includes the WMT-15 dataset, resulting in higher cross-BLEU compared to other language pairs.} 

\subsection{Test-set Size}

\begin{table}[!htp]\centering
\scalebox{0.9}{
\begin{tabular}{lrr}\toprule
\textbf{Size} &  \textbf{Bleu} & \textbf{Prism} \\\midrule
 100 & 0.735 & 0.720 \\
 200 & 0.783 & 0.771 \\
 400 & 0.804 & 0.784 \\
 800 & 0.828 & 0.807 \\
\bottomrule
\end{tabular}}
\caption{Average correlations versus test-set size for the language pairs from Figure~\ref{fig:cross-bleu}.} \label{tab:corr-vs-size}
\end{table}

In principle, a major advantage of reference-free evaluation is that it can make use of arbitrarily large test sets, being constrained only by the amount of source-language data in the domain of interest. We hypothesize that this will improve metric performance by reducing sampling error. To test this hypothesis in the absence of larger human-scored test sets for WMT19, we sampled subsets of various sizes and measured average correlation. As shown in Table~\ref{tab:corr-vs-size}, we observe a steady increase with test-size size. This provides persuasive, though not definitive, evidence that test sets beyond the scale of WMT19 would yield further improvements in accuracy for both metrics, a setting that would be more feasible for Prism than BLEU. Full curves are plotted in  Figure~\ref{fig:corr-vs-size} (See Appendix~\ref{sec:corr-vs-size}).



\section{Conclusion}

In this paper, we have shed some light on the remarkable finding by \prism\ that a multilingual model trained on a large (but not enormous) general-domain corpus can be highly effective as an MT metric when used to score the outputs of other MT systems in the absence of reference translations. By scaling up the model and making small adjustments to tagging and scoring, we improve over the original results and achieve approximate parity with BLEU in correlation with human judgment on WMT19 data. We argue that this metric is a useful complement to reference-based metrics---including ones that are significantly more powerful than BLEU---due to its flexibility; and we provide evidence that scoring reliability can be further improved by using larger source-side-only test sets.

We find that the major determinant of success in peer evaluation is the quality of the evaluating model. However, there is no hard requirement that it be better than the models under evaluation: surprisingly, it can correctly rank models that outperform it on average. If we abstract away from quality, performance does not appear to be highly sensitive to the domain or the multilingual versus bilingual nature of the training corpus. Taken together, these results have the important practical implication that a single multilingual system such as ours could be broadly applicable for evaluating systems in a large number of language pairs (706 in our case), at different quality levels, and across a wide range of domains. In future work, we look forward to probing these results further, and determining whether alternative architectures or loss functions might be valuable in specializing an MT model for evaluating its peers.

\section*{Acknowledgments}

We thank Julia Kreutzer, Ciprian Chelba, Aditya Siddhant, and the anonymous reviewers for their helpful and constructive comments.

\clearpage
\bibliography{refs}
\bibliographystyle{acl_natbib}

\appendix
\newpage

\onecolumn
\section{Outlier Systems}
\label{sec:outliers}

\begin{table}[!htp]\centering
\scalebox{0.75}{
\begin{tabular}{ll}\toprule
 \textbf{lang} &  \textbf{Outliers} \\\midrule
de-cs & CAiRE.6949 \\
de-en & online-X.0 \\
de-fr & - \\
en-cs & - \\
en-de & online-X.0, en\_de\_task.6790 \\
en-fi & apertium-fin-eng-unconstrained-en-fi.6448 \\
en-gu & - \\
en-kk & NICT.6550, DBMS-KU\_ENKK.6730 \\
en-lt & - \\
en-ru & NICT.6563 \\
en-zh & - \\
fi-en & - \\
fr-de & MSRA.MADL.6893, eTranslation.6262, online-X.0 \\
gu-en & Ju\_Saarland.6525 \\
kk-en & UMD.6736, DBMS-KU\_KKEN.6726 \\
lt-en & online-X.0 \\
ru-en & NICT.6561\\
zh-en & online-X.0, Apprentice-c.6706 \\
\bottomrule
\end{tabular}}
\caption{Outlier systems using MAD filtering in WMT19.} \label{tab:outliers}
\end{table}

\section{WMT 2019 System-Level results for all language pairs}
\label{sec:results-full}

\begin{table*}[!htp] \centering
\scalebox{0.80}{
\begin{tabular}{lr|rrrrrrrr}\toprule
\textbf{Metric} & Avg & en-cs & en-de & en-fi & en-gu & en-kk & en-lt & en-ru & en-zh \\\midrule
BLEU & 0.911 & 0.994 &	0.806 &	0.939 & 0.737 & 	0.575 &	0.986 &	0.946 &	0.802 \\
CHRF & 0.933 &  0.983&	0.871 &	0.964 &	0.843 & 0.829 &	0.969 &	0.989 &	0.799 \\
\midrule
UNI+ & 0.808 & -&- &- &- & -& -& 0.746 &-  \\
Yisi-2 & 0.487 &0.324  &  - & 0.478 & 0.314 & 0.685 & 0.055 & 0.134 & -0.097  \\
Prism & {\bf 0.861} &  0.865 &	{\bf 0.754} &	0.858 & {\bf 0.444} &	0.789 &	0.908 &	{\bf 0.903} &	{\bf 0.793} \\
\midrule
Prism-trg2xx & 0.853 & 0.867 &	0.717 &	0.876 & 0.365 &	0.811 &	0.936 &	0.902 &	0.778  \\
Prism-src2xx & 0.858 & {\bf 0.871} &	0.730 &	{\bf 0.878}&  0.400& 	{\bf 0.813} &	{\bf 0.939} &	0.901 &	0.789 \\
\bottomrule
\end{tabular}}

\vspace{0.2cm}
\scalebox{0.80}{
\begin{tabular}{lrrrrrrr|rrr}\toprule
\textbf{Metric} & de-en&  fi-en & gu-en & kk-en & lt-en & ru-en & zh-en   & de-cs & de-fr & fr-de \\\midrule
BLEU & 0.794 & 0.985 & 0.975 &  0.912 &	0.967 &	0.812	& 0.808& 0.743 & 0.891 & 0.846\\
CHRF & 0.852 & 0.991 & 0.946 & 0.836 &	0.930 &	0.877 &	0.831& 0.981 & 0.957 & 0.833\\
\midrule
UNI+ & 0.805 & 0.924 & -& -& -& 0.669 &- & - & -& -\\
Yisi-2 & 0.612 & 0.642 & 0.820 & 0.662 & 0.346 & 0.708 & 0.622&  0.122 & 0.721 & 0.62 \\
Prism & {\bf 0.829} &	0.941 &	{\bf 0.915} &	{\bf 0.724}&	0.985 &	{\bf 0.769} &	0.826& 0.987 & {\bf 0.889} & {\bf 0.269}\\
\midrule
Prism-trg2xx & 0.798 & {\bf 0.943} & 0.911 & 	0.683 &	{\bf 0.979} &	0.752 &	{\bf 0.830} & 0.989 & 0.882 & 0.212\\
Prism-src2xx &  0.796 &  0.942 & 0.893 & 0.709 &	0.978 &	0.739 &	0.828& {\bf 0.991} &  0.882 & 0.203\\
\bottomrule
\end{tabular}}
\caption{Baseline results. All numbers are system-level correlations. {\em Avg} gives averages over all language pairs.} \label{tab:baselines}
\end{table*}

\begin{table*}[!htp]\centering
\scalebox{0.80}{
\begin{tabular}{lr|rrrrrrrr}\toprule
\textbf{Data} & Avg & en-cs & en-de & en-fi & en-gu & en-kk & en-lt & en-ru & en-zh  \\\midrule
Prism-39 & 0.858 &  {\bf 0.871} &	0.730 & 0.878 &  0.400 & 	0.813 &	0.939 &	{\bf 0.901} &	0.789 \\
WMT-15 & 0.840 & 0.825 &	0.530 &	0.815 &	0.423 &	0.909 &	0.914 &	0.845 &	0.774 	 \\
Prism-13 & 0.863 & 0.869 &	{\bf \underline{0.779}} & { \bf 0.891} & 0.379	& 0.820& 0.925 &	{\bf 0.901} &	0.795\\
Prism-39 + WMT-15 & \underline{0.867} & 0.862 &	0.653 &	0.854 &  0.446 &	0.860 & 	0.932 &	0.880 & 	0.789 \\
\midrule
\textit{Adding monolingual data} \\
Prism-39 + Wiki-39 & 0.832 & 0.839 &	0.649 &	0.854 &	0.446 &	0.823	&0.917 &	0.877 &	0.757 \\
WMT-15 + WMT-15-Mono & {\bf 0.869} & 0.855	 &0.646 &	0.826 &	{\bf \underline{0.848}} &	{\bf 0.913} &	0.940 &	0.867 &	{\bf 0.793}\\
Prism-39 + WMT-15-Mono & 0.851 & 0.869 &	\underline{0.766} &	0.871 &	0.487 &	0.850 &	{\bf 0.945} &	0.892 &	0.754 \\
\midrule
Bilingual Models &  &  &0.726 & & & & 0.695 & 0.867 & 0.769\\
\bottomrule
\end{tabular}}

\vspace{0.2cm}
\scalebox{0.80}{
\begin{tabular}{lrrrrrrr|rrr}\toprule
\textbf{Data} & de-en&  fi-en & gu-en & kk-en & lt-en & ru-en & zh-en & de-cs & de-fr & fr-de \\\midrule
Prism-39 &  0.796 &  0.942 & 0.893 & 0.709 &	0.978 &	0.739 &	0.828& 0.991 &  0.882 & 0.203\\
WMT-15 &  0.815 & 0.954 & 0.918 & 0.509 &	0.986 &	0.841 &	0.835& 0.970 & 0.851 & 0.116\\
Prism-13 & {\bf \underline{0.845}} & 0.943 & 0.911 & {\bf 0.745} &	0.983 &	0.749 &	{\bf 0.836} & 0.985 & 0.863 & 0.124\\
Prism-39 + WMT-15 &  \underline{0.802} & 0.950 &  0.921 & 0.734 &	\underline{0.986} &	{\bf \underline{0.851}} &	0.810& {\bf 0.993} & 0.881 & 0.170\\
\midrule
\textit{Adding monolingual data} \\
Prism-39 + Wiki-39 &0.742 & 0.934 & 0.907 &	0.684 &	0.975 &	0.680 & 	0.836	& 0.982 & 0.822 & 0.117\\
WMT-15 + WMT-15-Mono & \underline{0.842} & {\bf \underline{0.956}} & {\bf \underline{0.976}} &	0.584 &	{\bf 0.987} &	0.838	 & 0.824 & 0.903 & {\bf 0.890} & {\bf 0.238} \\
Prism-39 + WMT-15-Mono &  0.803 & 0.944 & \underline{0.952} &	0.680 &	0.952	& 0.722 &	0.754& 0.972 & 0.875 & 0.233 \\
\midrule
Bilingual Models & 0.801 & & & & 0.862 & 0.650 & 0.826 & & & \\
\bottomrule
\end{tabular}
}
\caption{Effect of training data. Significant improvement over baseline ``Prism-39'' systems are underlined.} \label{tab:corpora}
\end{table*}

\begin{table*}[!htp]\centering
\scalebox{0.80}{
\begin{tabular}{lr|rrrrrrrr}\toprule
\textbf{Metric} & Avg & en-cs & en-de & en-fi & en-gu & en-kk & en-lt & en-ru & en-zh \\\midrule
BLEU & {\bf 0.911} & 
{\bf 0.994} &	0.806 &	{\bf 0.939} & 0.737 & 	0.575 &	{\bf 0.986} &	{\bf 0.946} &	0.802 \\
Big & 0.808 & 0.791	&0.541 &	0.833 &	0.381 &	0.785 &	0.898 &	0.861 &	0.698\\
Prism & 0.858 &  0.871 &	0.730 & 0.878 &  0.400 & 	0.813 &	0.939 &	0.901 &	0.789\\
Massive & 0.883 & 0.900 & {\bf 0.819} & 0.899	 & 0.423&	{\bf 0.820} &	0.953&	0.923&	{\bf 0.819} \\
\bottomrule
\end{tabular}}

\vspace{0.2cm}
\scalebox{0.80}{
\begin{tabular}{lrrrrrrr|rrr}\toprule
\textbf{Metric} & de-en&  fi-en & gu-en & kk-en & lt-en & ru-en & zh-en   & de-cs & de-fr & fr-de \\\midrule
BLEU & 0.794 & {\bf 0.985} & {\bf 0.975} &  {\bf 0.912} &	0.967 &	{\bf 0.812}	& 0.808& 0.743 & 0.891 & {\bf 0.846}\\
Big & 0.702 & 0.926 &	0.891 &	0.649 &	0.970 &	0.640 &	0.819 & 0.989 & 0.827 & 0.040\\
Prism &  0.796 &  0.942 & 0.893 & 0.709 &	0.978 &	0.739 &	0.828& 0.991 &  0.882 & 0.203 \\
Massive & {\bf 0.840} &  0.948 &	0.906 &	0.751 &	{\bf 0.981} &	0.789 &	{\bf 0.834} & 
{\bf 0.991} & {\bf 0.906} & 0.301\\
\bottomrule
\end{tabular}}
\caption{Effect of Model capacity.
} \label{tab:capacity}
\end{table*}

\begin{table*}[!htp]\centering
\scalebox{0.80}{
\begin{tabular}{lr|rrrrrrrr}\toprule
\textbf{Aggregation Method} & Avg & en-cs & en-de & en-fi & en-gu & en-kk & en-lt & en-ru & en-zh \\\midrule
Mean & 0.883 & 0.900 & 0.819 & 0.899	 & 0.423&	0.820 &	0.953&	0.923&	0.819 \\
Std-dev &  0.882 &0.913 &	0.778 &	0.900&	0.448 &	0.755&	0.950 &	0.929 &	0.780 \\
Median & 0.870 & 0.849&	{\bf 0.868}	&0.884 &	0.408 &	0.862 &	0.945 &	0.908 &	0.849\\
Min  & 0.872 & 0.925 &	0.765 &	0.907 &	 0.489	&0.623 &	0.945&	0.939 &	0.722\\
\midrule
MC-dropout (Mean) & 0.877 &  \underline{0.936} &	0.826 &	 0.904 &	0.432 &	0.699&	0.929 &	{\bf 0.940} &	{\bf 0.881}	\\
MC-dropout (Std-dev)  & 0.855 & 0.890 &	0.800 &	0.894 &	0.417 &	0.809 &	0.947 &	0.920 &	0.803\\
SP-Norm (Mean) &  0.884 & 0.903 &	0.814 &	0.895 &	0.407 &	\underline{\bf 0.872} &	 0.949 &	0.921 &	\underline{0.839}\\
\midrule
Confidence threshold & {\bf 0.886} & \underline{\bf 0.941} & 0.828 & \underline{\bf 0.966} & {\bf 0.569} & 0.696 & \underline{\bf 0.987} & {\bf 0.940} & 0.774 \\
\bottomrule
\end{tabular}}

\vspace{0.2cm}
\scalebox{0.80}{
\begin{tabular}{lrrrrrrr|rrr}\toprule
\textbf{Aggregation Method} & de-en&  fi-en & gu-en & kk-en & lt-en & ru-en & zh-en   & de-cs & de-fr & fr-de \\\midrule
Mean &  {\bf  0.840} &  0.948 &	0.906 &	0.751 &	0.981 &	0.789 &	0.834 & 0.991 & 0.906 & 0.301\\
Std-dev & 0.831 & {\bf 0.970} &	{\bf 0.946} &	0.712 &	0.987 &	0.769 &	0.809 & 0.989 & 0.907 & 0.276\\
Median & 0.851 & 0.895 &	0.885 &	0.761 &	0.970 &	0.788 &	{\bf 0.870} & 0.973 & 0.888 & 0.341\\
Min  & 0.829 & 0.973 &	0.902 &	0.644 &	{\bf 0.990} &	0.757	&0.794 & 0.978 & 0.900 & 0.280\\
\midrule
MC-dropout (Mean) &  0.813 & 0.956	&	0.485 &	0.189 &	0.941 &	{\bf 0.834} &	0.821 & 0.974 & 0.923 & 0.330\\
MC-dropout (Std-dev)  & 0.837 & 0.946 &		0.903 &	0.743 &	0.981 &	0.786 &	0.837 & {\bf 0.992} & 0.901 & 0.286 \\
SP-Norm (Mean) & 0.834 & 0.948&	0.902 &	{\bf 0.795} &	0.980 &	0.801 &	0.833 & 0.990 & 0.906 & 0.318 \\
\midrule
Confidence threshold & 0.823 & 0.930 & 0.906 & 0.670 & 0.970 & 0.716 & 0.765  & 0.971 &  {\bf 0.935} & {\bf 0.386} \\
\bottomrule
\end{tabular}}
\caption{Scoring methods. Significant improvements over baseline {\em Mean} systems are underlined.} \label{tab:alternate}
\end{table*}

\newpage

\section{Correlation versus test-set size}
\label{sec:corr-vs-size}
\begin{figure*}[htp]
    \centering
    \includegraphics[scale=0.5]{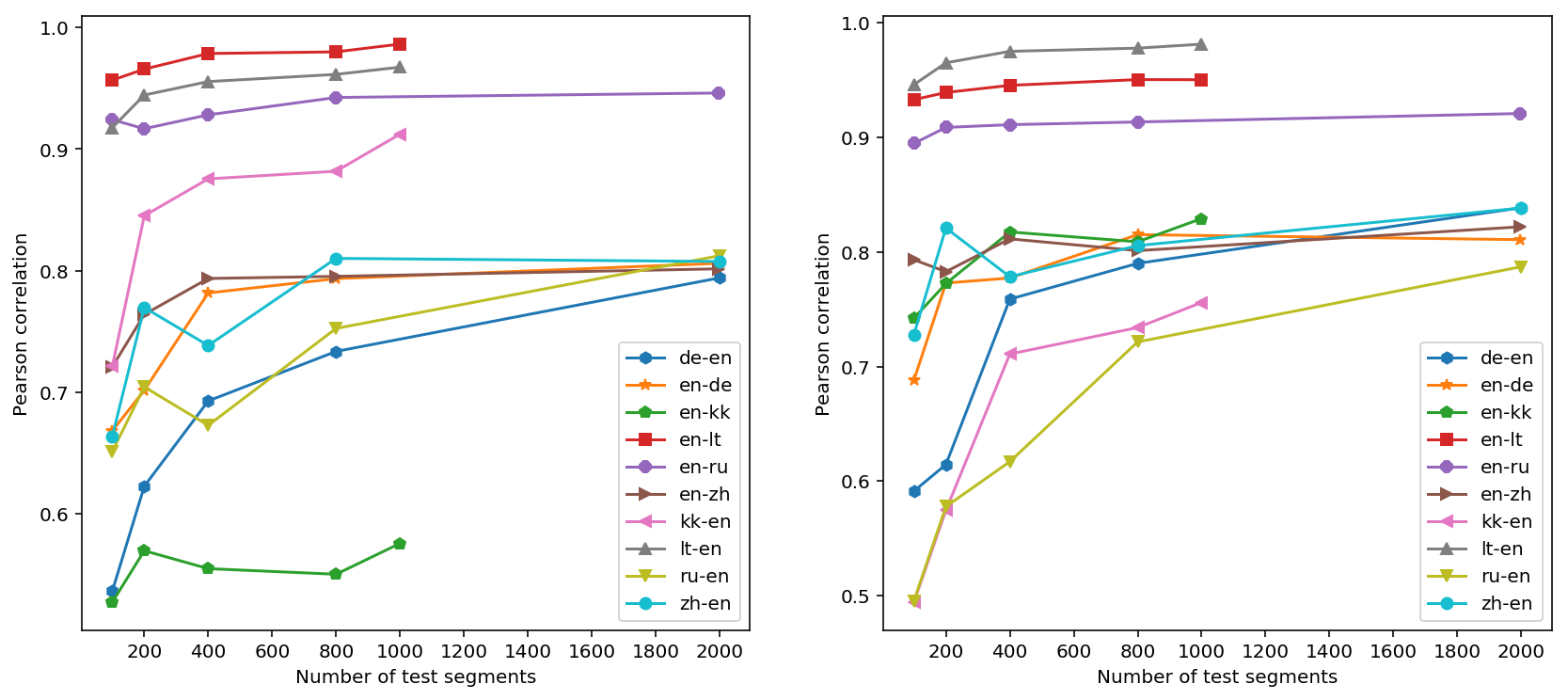}
    \caption{Correlation as test-size size increases, for BLEU (left panel) and Prism (right panel). Each point is the average correlation over 10 random draws of subsets of the given size.}
    \label{fig:corr-vs-size}
\end{figure*}

\end{document}